\documentclass[10pt,twocolumn,letterpaper]{article}

\usepackage[pagenumbers]{cvpr} 

\usepackage[accsupp]{axessibility}  

\usepackage{graphicx}
\usepackage{amsmath}
\usepackage{amssymb}
\usepackage{booktabs}
\usepackage{enumerate}
\usepackage{multirow}
\usepackage{float}

\usepackage{times}
\usepackage{epsfig}
\usepackage{graphicx}
\usepackage{amsmath}
\usepackage{amssymb}
\usepackage{enumerate}
\usepackage{booktabs}
\usepackage{multirow}
\usepackage[table,xcdraw]{xcolor}
\usepackage{booktabs}
\usepackage{multirow}
\usepackage[normalem]{ulem}
\useunder{\uline}{\ul}{}

\usepackage{wrapfig}

\usepackage{times}
\usepackage{epsfig}
\usepackage{graphicx}
\usepackage{amsmath}
\usepackage{amssymb}
\usepackage{tabulary}
\usepackage{multirow}
\usepackage{silence}
\usepackage{caption}
\usepackage{xfrac}
\usepackage{float}
\usepackage[pagebackref=true,breaklinks=true,letterpaper=true,colorlinks,bookmarks=false]{hyperref}

\usepackage{algorithm}
\usepackage{algpseudocode}
\usepackage[algo2e]{algorithm2e}

\makeatletter
\renewcommand\paragraph{\@startsection{paragraph}{4}{\z@}%
                                    {0.5ex \@plus1ex \@minus.1ex}%
                                    {-1em}%
                                    {\normalfont\normalsize\bfseries\maybe@addperiod}}
\newcommand{\maybe@addperiod}[1]{%
  #1\@addpunct{.}%
}
\newcommand{\vspaceaftersection}{-0.18cm}
\newcommand{\vspacebeforesection}{-0.18cm}

\newcommand{\vspaceafterequation}{-0.08cm}
\newcommand{\vspacebeforeequation}{-0.08cm}

\newcommand{\latent}{\Theta}
\newcommand{\latentcenter}{\Theta_{\text{c}}}
\newcommand{\latentR}{\Theta_{\text{R}}}
\newcommand{\latentinv}{\Theta_{\text{inv}}}
\newcommand{\latentscale}{\Theta_{\text{s}}}
\newcommand{\pc}{P}
\newcommand{\pcCentroid}{\bar{P}}

\newcommand{\VNL}{\text{VN}}
\newcommand{\VNLN}{\widehat{\text{VN}}}
\makeatother

\usepackage[capitalize]{cleveref}
\crefname{section}{Sec.}{Secs.}
\Crefname{section}{Section}{Sections}
\Crefname{table}{Table}{Tables}
\crefname{table}{Tab.}{Tabs.}

\def\cvprPaperID{538}
\def\confName{CVPR}
\def\confYear{2023}
\begin{document}
\title{\emph{EFEM}: \emph{E}quivariant Neural \emph{F}ield \emph{E}xpectation \emph{M}aximization \\ for 3D Object Segmentation Without Scene Supervision}

\author{
        Jiahui Lei\textsuperscript{1} \quad Congyue Deng\textsuperscript{2} \quad Karl Schmeckpeper\textsuperscript{1} \quad  Leonidas Guibas\textsuperscript{2} \quad Kostas Daniilidis\textsuperscript{1}\\
        $^1$ University of Pennsylvania \qquad
        $^2$ Stanford University\\
        {\tt\small \{leijh, karls, kostas\}@cis.upenn.edu, \{congyue, guibas\}@cs.stanford.edu} 
    }

\maketitle\vspace{-0.5em}

\begin{abstract}
We introduce Equivariant Neural Field Expectation Maximization (\textbf{EFEM}), a simple, effective, and robust geometric algorithm that can segment objects in 3D scenes without annotations or training on scenes.
We achieve such unsupervised segmentation by exploiting single object shape priors. 
We make two novel steps in that direction. 
First, we introduce equivariant shape representations to this problem to eliminate the complexity induced by the variation in object configuration. 
Second, we propose a novel EM algorithm that can iteratively refine segmentation masks using the equivariant shape prior. 
We collect a novel real dataset \textit{Chairs and Mugs} that contains various object configurations and novel scenes in order to verify the effectiveness and robustness of our method. Experimental results demonstrate that our method achieves consistent and robust performance across different scenes where the (weakly) supervised methods may fail. Code and data available at \small{\url{https://www.cis.upenn.edu/~leijh/projects/efem}}
\end{abstract}
\vspace{-1.5em}
\section{Introduction}
\vspace{\vspaceaftersection}
\label{sec:intro}

Learning how to decompose 3D scenes into object instances is a fundamental problem in visual perception systems.
Past developments in 3D computer vision have made huge strides on this problem by training neural networks on 3D scene datasets with segmentation masks~\cite{schult2022mask3d,vu2022softgroup,wu20223d}. However, these works heavily rely on large labeled datasets~\cite{dai2017scannet,Matterport3D} that require laborious 3D annotation based on special expertise.
Few recent papers alleviate this problem by reducing the need to either sparse point labeling~\cite{hou2021exploring,tang2022learning} or bounding boxes~\cite{chibane2022box2mask}.

In this work, we follow an object-centric approach inspired by the Gestalt school of perception that captures an object as a whole shape \cite{koffka35,palmer99} invariant to its pose and scale \cite{kendall1989survey}. 
A holistic approach builds up a prior for each object category, that then enables object recognition in different complex scenes with varying configurations. Directly learning object-centric priors instead of analyzing each 3D scene inspires a more efficient way of learning instance segmentation: both a mug on the table and a mug in the dishwasher are mugs, and one does not have to learn to segment out a mug in all possible environmental contexts if we have a unified shape concept for mugs.
Such object-centric recognition facilitates a robust scene analysis for autonomous systems in many interactive real-world environments with a diversity of object configurations: Imagine a scenario where a robot is doing the dishes in the kitchen. Dirty bowls are piled in the sink and the robot is cleaning them and placing them into a cabinet. Objects of the same category appear in the scene repeatedly under different configurations (piles, neat lines in the cabinet). What is even more challenging is that even within this one single task (doing dishes) the scene configuration can drastically change when objects are moved.
We show that such scenarios cannot be addressed by the state-of-the-art strongly or weakly supervised methods that struggle under such scene configuration variations.

\begin{figure}
    \centering
    \includegraphics[width=\linewidth]{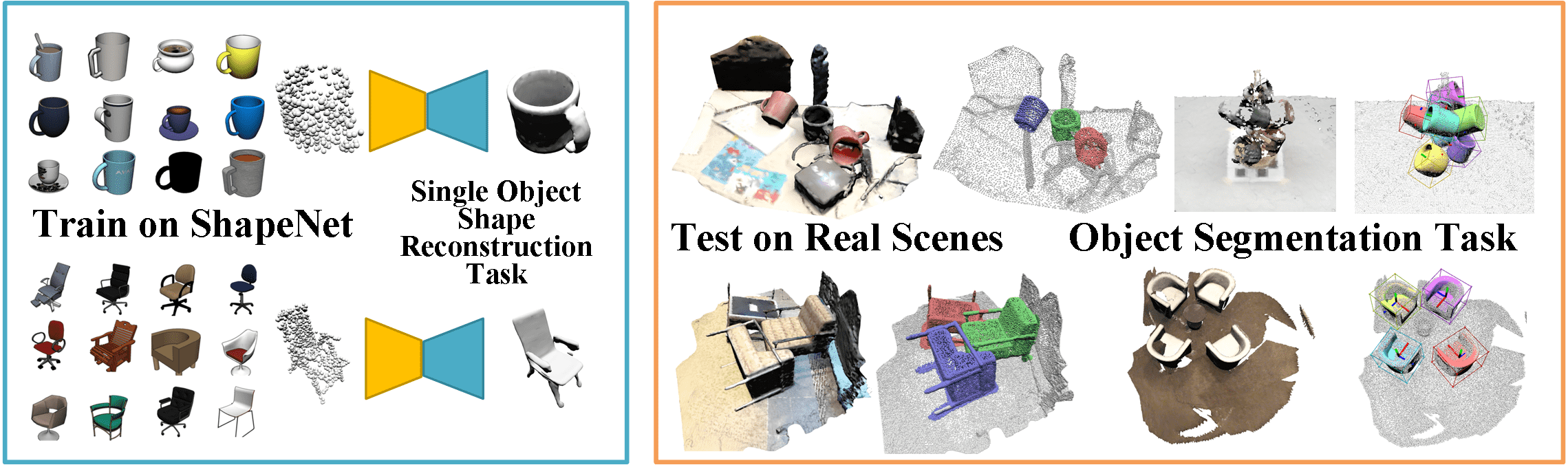}
    \caption{We present EFEM, an unsupervised 3D object segmentation method applicable to real-world scenes (results on the right) by only training on ShapeNet single object reconstruction.}
    \label{fig:ser}
\end{figure}

In this paper, we introduce a method that can segment 3D object instances from 3D static scenes by learning priors of single object shapes (ShapeNet~\cite{chang2015shapenet}) without using any scene-level labels.
Two main challenges arise when we remove the scene-level annotation. First, objects in the scene can have a different position, rotation, and scale than the canonical poses where the single object shapes were trained. Second, the shape encoder which is trained on object-level input cannot be directly applied to the scene observations unless the object masks are known.
We address the first challenge by introducing equivariance to this problem. By learning a shape prior that is equivariant to the similitude group SIM(3), the composition of a rotation, a translation, and a uniform scaling in 3D (Sec.~\ref{sec:shape_prior}), we address the complexity induced by the SIM(3) composition of objects. For the second challenge, we introduce a simple and effective iterative algorithm, Equivariant neural Field Expectation Maximization (\textbf{EFEM}), that refines the object segmentation mask, by alternately iterating between mask updating and shape reconstruction (Sec.~\ref{sec:single_alg}).
The above two steps enable us to directly exploit the learned single instance shape prior to perform segmentation in real-world scenes. We collected and annotated a novel real-world test set (240 scenes) (Sec.~\ref{sec:exp_real}) that contains diverse object configurations and novel scenes to evaluate the generalizability and robustness to novel object instances and object configuration changes. Experiments on both synthetic data (Sec.~\ref{sec:exp_syn}) and our novel real dataset (Sec.~\ref{sec:exp_real}) give us an insight to the effectiveness of the method. 
Compared to weakly supervised methods, 
when the testing scene setup is similar to the training setup, our method has a small performance gap to the (weakly) supervised baselines. However, when the testing scenes are drawn from novel object configurations, our method consistently outperforms the (weakly) supervised baselines.

Our paper makes the following novel contributions to the 3D scene segmentation problem:
(1) a simple and effective iterative EM algorithm that can segment objects from the scenes using only single object shape priors. 
(2) addressing the diversity of object composition in 3D scenes by combining representations equivariant to rotation, translation, and scaling of the objects.
(3) an unsupervised pipeline for 3D instance segmentation that works in real-world data and can generalize to novel setups. 
(4) a novel real-world test set \textbf{Chairs and Mugs} that contains diverse object configurations and scenes.

\vspace{\vspacebeforesection}
\section{Related Work}
\label{sec:related}
\vspace{\vspaceaftersection}

\paragraph{3D Instance segmentation}

Point cloud instance segmentation has been a long-existing challenge in 3D vision even before the existence of deep learning. 
Early interests have been focusing on 3D object retrieval in scenes where RANSAC and generalized Hough voting were the most prominent paradigms~\cite{schnabel2007efficient}.
In recent years, with the proliferation of large synthetic \cite{wu2018building, fu20213d} or real \cite{hua2016scenenn, Matterport3D, dai2017scannet, hou2019sis} datasets with rich annotations, learning-based methods have shown great success on this task.

\paragraph{Supervised methods}
Most supervised methods fall into two categories: top-down methods that locate objects first and then predict the refined segmentation mask \cite{yang2019learning, hou20193d}, and bottom-up methods learning to group points into object proposals.
Regarding the grouping approach, a variety of algorithms have been explored, such as point-pair similarity matrices \cite{wang2018sgpn, zhang2021point}, mean-shift clustering \cite{lahoud20193d}, graph-based grouping \cite{han2020occuseg}, cluster growing \cite{jiang2020pointgroup}, hierarchical aggregation \cite{chen2021hierarchical, liang2021instance}, or adversarial methods \cite{yi2019gspn}. Later works \cite{vu2022softgroup, vu2022softgroup++} combine top-down and bottom-up approaches and achieve impressive results. Recently, transformers and attention mechanisms have also been introduced to this problem~\cite{schult2022mask3d}.
\cite{nie2021rfd} also incorporates neural implicit representations and simultaneously performs segmentation and shape reconstruction.
Despite their requirements for laborious data annotation, we will also show that supervised methods heavily rely on the correlation between training and test scenes, and even the state-of-the-art supervised methods struggle to generalize to novel scene configurations or to changes in background patterns.

\paragraph{Unsupervised or weakly supervised methods}
Many attempts have been made to learn instance segmentation with limited annotations.
A number of works leverage pre-extracted features from scenes via representation learning such as graph attentions \cite{lee2022gaia} or contrastive learning \cite{xie2020pointcontrast, hou2021exploring, yang2021unsupervised,chu2022twist} to facilitate weakly-supervised training with fewer point labels.
Other approaches directly propagate sparse annotations to dense point labels by learning point affinity graphs \cite{tang2022learning} or by bounding box voting \cite{chibane2022box2mask}.
These methods are usually the most scalable to large scenes, but their point features are scene-dependent, and we will show that similarly to fully-supervised learning methods, weakly-supervised methods have difficulties in generalizing to scenes with different configurations.
When confronting dynamic scenes, one can exploit temporal self-consistency  \cite{song2022ogc} or scene pair constraints \cite{huang2021multibodysync, yang2021unsupervised} but such constraints  introduce additional assumptions about the scenes.
Most related to our work are the retrieval-based methods leveraging object priors \cite{li2015database, xie2022improved}. Traditional retrieval methods are either limited to one given object template \cite{xie2022improved} or need to solve a discrete combinatorial optimization problem searching for the target template in the object category \cite{li2015database}.
We resolve these issues by learning an implicit shape prior which can be optimized continuously in the feature space.

\paragraph{Implicit object priors}

First introduced in \cite{deepsdf, imnet, onet}, neural implicit representations (also known as neural fields~\cite{srinathsurvey}) parameterize 3D shapes as level sets of neural networks. They not only show strong capabilities of capturing geometric details within limited capacity \cite{sitzmann2020implicit, tancik2020fourfeat, peng2020convolutional, jiang2020local} but also avoid shape discretization that introduces sampling noises.
When representing a collection of objects with a shared implicit network, its bottleneck layer naturally forms a latent embedding of the objects, which can serve as a shape prior for many downstream tasks such as shape generation \cite{imnet, ibing20213d, liu2022towards}, reconstruction \cite{lin2020sdfsrn}, completion \cite{deepsdf, onet, chibane2020implicit}, computing correspondences \cite{kohli2020semantic, simeonov2022neural, liu2020learning, lei2022cadex}, and part decomposition \cite{chen2019bae}. We refer the readers to~\cite{srinathsurvey} for a comprehensive review.
Our method takes advantage of the recent work in neural fields to learn a deep shape prior.

\paragraph{Equivariant point cloud networks}
Equivariant networks are designed to preserve transformation coherence between the input and latent representations. With well-developed theories~\cite{cohen2018general,kondor2018generalization,  weiler2021coordinate,aronsson2022homogeneous,xu2022unified}, equivariant networks have a variety of designs on pointclouds~\cite{deng2021vector,thomas2018tensor, fuchs2020se,poulenard2021functional,assaad2022vn,katzir2022shape}, which benefits many downstream tasks such as robotic applications~\cite{simeonov2022neural,weng2022neural,higuera2022neural,xue2022useek,fu2022robust,ryu2022equivariant}, 3D reconstruction~\cite{chen2021equivariant,chatzipantazis2022se}, and object pose estimation/canonicalization~\cite{zhu2022correspondence,pan2022so,li2021leveraging,lin2022coarse,sajnani2022_condor}.
Unlike these object-level works, we focus on leveraging equivariance object features in scene understandings.
Most related to us is \cite{yu2022rotationally} which also applies object equivariance to scenes. But they perform supervised 3D object bounding box detection while we predict dense instance masks.
In this work, we employ the vector neurons~\cite{deng2021vector,chen2021equivariant,assaad2022vn} to build our equivariant shape prior.

\vspace{\vspacebeforesection}
\section{Method}
\vspace{\vspaceaftersection}

\begin{figure*}[t!]
\centering
   \includegraphics[width=1.0\textwidth]{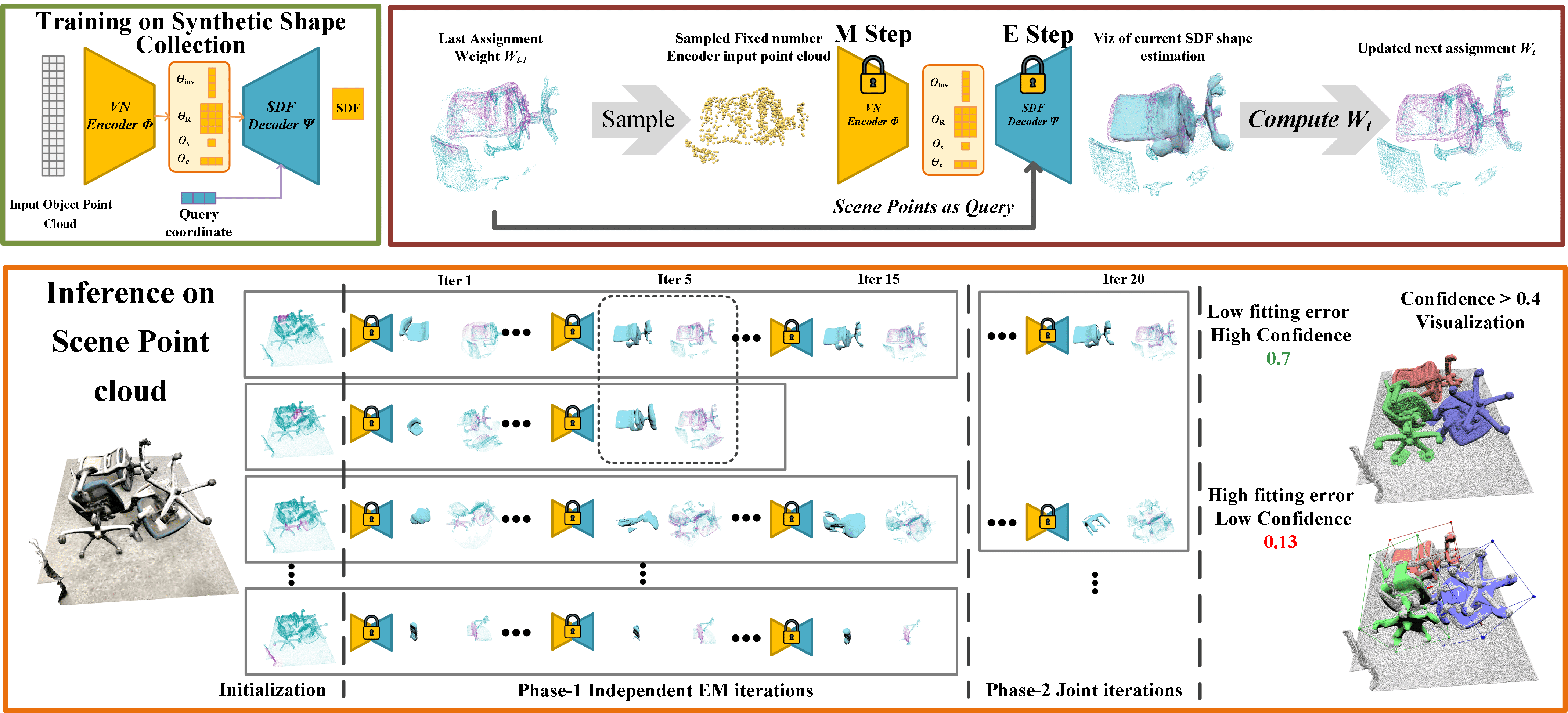}
   \\\vspace{-.7em}
    \caption{
    \textbf{Overview}: \textbf{Left top}: the single object SDF Encoder-Decoder (Sec.~\ref{sec:shape_prior}) is trained on the shape collection. Once trained, the network's weights are frozen.
    \textbf{Top right} Single EM step on scene observations (Sec.~\ref{sec:single_alg}): given the last estimation of the object mask $W_{t-1}$, a set of points is sampled from the full scene point cloud and then passed to the shape encoder $\Phi$ to produce the current estimation of the shape embedding $\latent$. Based on the $\latent$, all the scene points are queried again through the decoder $\Psi$ to generate a new assignment mask $W_t$.
    \textbf{Bottom} Object segmentation pipeline: starting from random crop initializations from the left, the above EM step is applied to each proposal (each row) in parallel to refine their masks. In the early steps (Phase-1 Sec.~\ref{sec:single_alg}), all the proposals run independently but in the second phase, multiple proposals can be jointly optimized (Sec.~\ref{sec:full_alg}). Note that we eliminate the duplicated (the second) or unreasonably sized (the fourth) proposals during optimization. Finally, the remaining proposals and their confidences are output. 
    \vspace{-1em}
    }
    \label{fig:method}
\end{figure*}

Now we introduce our method for unsupervised object segmentation in 3D scenes.
At the training stage, we learn an object-level shape prior (Sec.~\ref{sec:shape_prior}, Fig.~\ref{fig:method} top left) utilizing a collection of synthetic models from an object category (e.g. all chairs in ShapeNet~\cite{chang2015shapenet}).
At inference time, we are given a scene point cloud $X_{N \times 6}$ of $N$ points with coordinates as well as normal vectors with an unknown number of novel instances from the object category, and our task is to predict their instance segmentation masks.
We will introduce a simple and novel iterative algorithm for predicting instance segmentation masks,
starting from a single-object proposal phase (Sec.~\ref{sec:single_alg}, Fig.~\ref{fig:method} top right) 
and followed by a multiple-object joint proposal phase (Sec.~\ref{sec:full_alg}, Fig.~\ref{fig:method} bottom).
As by-products, our model also outputs implicit surface reconstructions, poses, and bounding boxes.

\vspace{\vspacebeforesection}
\subsection{SIM(3) Equivariant Shape Priors}
\label{sec:shape_prior}
\vspace{\vspaceaftersection}
Most synthetic datasets have their objects manually aligned to canonical poses and unit scales, yet the SIM(3) transformations (translations, rotations, and scales) must be considered when applying the shape priors to real-world scenes.
To this end, we constructed a SIM(3)-equivariant SDF encoder-decoder following the paradigms of prior work~\cite{onet,deng2021vector} (Fig.~\ref{fig:method} left top).

\paragraph{Point cloud encoder}
Given an object point cloud $\pc_{N_O\times 3}$ with $N_O$ points, it is first encoded by a SIM(3)-equivariant encoder $\Phi$ (yellow block Fig.~\ref{fig:method} left) constructed with Vector Neurons (VN)~\cite{deng2021vector,assaad2022vn,chen2022equivariant} into a latent embedding $\latent = \Phi(P)$ (orange block Fig.~\ref{fig:method} middle).
More concretely, the input point cloud $\pc$ is first subtracted by its centroid $\pcCentroid$ for translation equivariance, followed by a backbone network providing a global vector-channeled embedding $F$, which is scale- and rotation-equivariant and translation invariant. 
$F$ is then mapped to the shape implict code $\latent$ comprising four components $(\latentR, \latentinv, \latentcenter, \latentscale)$.
The backbone network is a rotation-equivariant VN Point-Transformer~\cite{assaad2022vn} with additional scale equivariance enforced by channel-wise normalizations as in \cite{chen2022equivariant}. We denote the modified linear layer with scale invariance \cite{chen2022equivariant} as $\VNLN$.
The four components of $\Theta$ are computed from $F$ with four heads separately:
\begin{enumerate} 
    \itemsep0em 
    \item A vector-channeled rotation equivariant latent code $\latentR=\VNLN_R(F)$.
    \item A scalar-channeled invariant latent code $\latentinv=\langle \VNLN_I(F), \latentR \rangle$ computed with inner product.
    \item A scalar $\latentscale$ by taking the average of $F$'s per-channel norm which explicitly encodes the object scale.
    \item A centroid correction vector predicting the offset between the centroid of points and the actual object center, which could be different due to the partiality and noises of the point could: $\latentcenter=\VNL_C(F) + \pcCentroid$, where $\VNL_C$ has $1$ output vector channel.
\end{enumerate}
Network architecture details can be found in the supplementary.
For any transformation $g=(s,R,t) \in \mathrm{SIM}(3)$ with scale $s$, rotation $R$, and translation $t$, its action on $\latent$ and the equivariance of the encoder $\Phi$ can be written as:
\begin{equation}
    \vspace{\vspacebeforeequation}
    g \circ \latent = 
    (\latentR R, \latentinv, s\latentcenter R+t, s \latentscale)
    = \Phi(s P R + t).
    \vspace{\vspaceafterequation}
\end{equation}

\paragraph{SDF decoder}
Give a query position $x \in \mathbb{R}^3$, its SDF value $\hat{v}(x)$ is predicted as
\begin{equation}
    \vspace{\vspacebeforeequation}
    \hat{v}(x) = \Psi(x; \latent) =\Psi(\latentinv, \langle \latentR, \Tilde{x}\rangle),
    \vspace{\vspaceafterequation}
\end{equation}
where $\Tilde{x}=(x-\latentcenter)/\latentscale$ is the canonicalized coordinate of $x$ with center $\latentcenter$ and scale $\latentscale$, and $\Psi$ is an MLP as in \cite{deng2021vector} that decodes the concatenation of the invariant feature $\latentinv$ and the channel-wise inner product between $\latentR$ and $\Tilde{x}$.

\paragraph{Training}
The implicit shape prior is trained with the standard L2 loss for the query points sampled around each object. 
To avoid arbitrary prediction of $\latentcenter$ and $\latentscale$ that may make training unstable in early epochs, we regularize $\latentcenter$ to stay around zero and $\latentscale$ to stay around one.
To help the network's generalization to real-world scenarios with partial observations, clutters, and sensor noises, we further augment the input object point clouds with partial depths and content-wise augmentations. 
Additional details of these augmentations are provided in the supplementary. 
We will next introduce how to exploit this learned shape-prior network which only takes instance-level inputs in scene-level point cloud segmentation.

\subsection{Iterative Algorithm for Single Proposal}
\label{sec:single_alg}
\begin{algorithm}[H]
 \KwData{ Scene Point Cloud $X_{N\times 6}$; trained encoder $\Phi$ and decoder $\Psi$ with frozen weight}
 Initialize $W_0$\;
 
 \While{not reach max step}{
  
  M-step: Sample $P_t$ by Eq.~\ref{eq:sample} from $W_{t-1}$ and forward encoder $\Phi$ to update $\latent_t$. \;
  
  E-step: Evaluate decoder $\Psi$ and update $W_t$ by Eq.~\ref{eq:w_update}.
 }

 Extract mesh and compute absolute pose as a byproduct.
 
 Compute confidence $C$ in Eq.~\ref{eq:confidence} and output mask.
  
 \caption{Single Proposal Estimation}
 \label{alg:single}
\end{algorithm}

As shown in Fig~\ref{fig:method}, the key component of our full algorithm is the single proposal processing (right top) since the full algorithm is constructed by many single proposals being processed in parallel.
Each proposal is represented by a soft assignment mask $W$ over all scene points, where the continuous value on each point $W[i]\in [0,1]$ indicates how likely this point belongs to the object that the proposal is representing.
In the early single proposal iteration steps (Phase-1), each proposal aims to fit its $W$ to one object independently. In later iterations, we optimize the proposals jointly, as discussed in Section~\ref{sec:full_alg}.

\paragraph{Initialization}
The single proposal algorithm starts from the initial assignment $W_0$ which is drawn from a random ball or cylinder cropping of the scene. The radius of the crop is set to be similar to the average size of the target class. All points inside the crop are set to have $W[i]=1.0$ while all other points are set to have $W[i]=0.0$.

\paragraph{M-Step: Estimating shape embeddings}
We treat the learned decoder with fixed weight as a parametric model of shape categories driven by the parameter $\latent$.
In each iteration of our algorithm, a new $\latent_t$ is estimated from the last assignment $W_{t-1}$.
Note that the learned shape prior encoder $\Phi$ only accepts a fixed number $N_O$ of points as input, which is always far less than the number of points in the scene observation. Therefore at each iteration, we first sample a fixed number $N_O$ of points from the scene point cloud $X_{N\times 6}$ based on the last assignment estimation $W_{t-1}$
\begin{equation}
    \vspace{\vspacebeforeequation}
    P_{t} = \mathcal M(W_{t-1}, X_{N\times 6}), \label{eq:sample}
    \vspace{\vspaceafterequation}
\end{equation}
where the sampling operations $\mathcal M$ contain two steps. 
First, a sample is drawn from a Bernoulli distribution with positive probability $W_{t-1}[i]$ to determine if the point $X[i]$ belongs to the object's foreground.
Second, for all points that are marked as foreground, we globally apply multinomial sampling based on their $W[i]$ to find $N_O$ sampled points. 
Finally, we produce the new shape estimation by passing the sampled point cloud through $\Phi$ so $\latent_t=\Phi(P_t)$. 
This method of updating the shape parameter can be interpreted as the M-step of an EM algorithm, which computes better distribution parameters based on the last assignment via weighted maximum likelihood. 

\paragraph{Fitting error}
Given the $\latent$ estimation, the fitting error of one observed point $\mathbf{x}=[x_\text{obs}, n_\text{obs}]$ ($x_\text{obs}$ denotes position and $n_\text{obs}$ denotes normal) in the scene point cloud $X_{N\times 6}$ is:
\begin{equation}
    \vspace{\vspacebeforeequation}
    e_{D}(\mathbf{x}, \latent) = |\Psi(x_\text{obs}; \latent)|
    \label{eq:error_dist}
    \vspace{\vspaceafterequation}
\end{equation}
\vspace{-1em}
\begin{equation}
    \vspace{\vspacebeforeequation}
    e_{N}(\mathbf{x}, \latent) = \text{acos} \left( \frac{n_\text{obs}^T \nabla_x \Psi(x_\text{obs}; \latent)}{\|\nabla_x \Psi(x_\text{obs}; \latent)\|} \right)
    \label{eq:error_normal}
    \vspace{\vspaceafterequation}
\end{equation}
\begin{equation}
    \vspace{\vspacebeforeequation}
    E(\mathbf{x}, \latent) = \alpha_D e_{D}(\mathbf{x}, \latent) + \alpha_N e_{N}(\mathbf{x}, \latent)
    \label{eq:error}
\end{equation}
which measures the observed point distance to the zero level set of the SDF and the normal consistency between the observed point and the decoded SDF gradient. 
The hyperparameters $\alpha_D$ and $\alpha_N$ control the importance of the distance and angle terms respectively.

\paragraph{E-Step: Updating point assignments}
After updating the shape parameters $\latent_t$, we update the assignment weight $W_t$ by querying the decoder $\Psi$ for all the points in the scene point cloud $X$ and computing the error in Eq.~\ref{eq:error}. The new assignment is designed to be updated as:
\begin{equation}
    \vspace{\vspacebeforeequation}
    W_t[i] = \frac{e^{-E(X[i], \latent_t)}}{e^{-E(X[i], \latent_t)} + \Omega},\label{eq:w_update}
    \vspace{\vspaceafterequation}
\end{equation}
where $\Omega$ is a constant hyperparameter that gives every point some probability to be in the background. This step can be interpreted as the E-step in an EM algorithm.

\paragraph{Termination and confidence score}
When the initialization is not near an object instance of the target class, the shape prior tends to never fit the input observations, so that the error $E$ is large everywhere and the weight $W$ is always small. 
We terminate the proposals that have less than a predefined threshold of small-error points at each iteration.
After the last iteration, we use Marching Cubes to extract a mesh $(\mathcal{V}, \mathcal{E})$ for each proposal and the mesh serves as the byproduct of our method. 
Our algorithm also produces a pose estimation with respect to the training shape collection via Procrustes registration~\cite{zhu2022correspondence} on $\latent_{SO3}$ between the observed objects and the member of the training set with the most similar $\latent$ (see Supp.).  
We also remove proposals with meshes outside a predefined reasonable range of scales.

One advantage of utilizing our shape prior is that we can explicitly compute the confidence score from the shape reconstruction.
Specifically, we compute two scores by measuring the fitting errors: 
(1.) \textbf{Observation fitting score}: a good fitting should have all the encoder input points located on the decoded SDF zero level set. 
We measure the proportion of the encoder input point cloud which has a small distance and angle error:
\begin{equation}
    \vspace{\vspacebeforeequation}
    S_1 = \frac{1}{N_O} \left|\left\{\mathbf{x}\in P_T
    \left| \begin{array}{c}
         e_{D}(\mathbf{x}, \latent)<\delta_D,  \\
         e_{N}(\mathbf{x}, \latent) < \delta_N 
    \end{array} \right.
    \right\}\right|,
    \label{eq:fitting_score}
    \vspace{\vspaceafterequation}
\end{equation}
where the $\delta_D,\delta_N$ are the thresholds.
(2.) \textbf{Reconstruction coverage score}: 
Since we recognize objects by their shapes, we should be less confident in detections where the observed points only cover a small portion of the extracted mesh.
For every vertex $\mathbf{v}=[x_\text{recon}, n_\text{recon}]$ from the extracted mesh, 
we find its nearest neighbour $\mathbf{x}=[x_\text{obs}, n_\text{obs}]$ in the observed scene point cloud $X_{N\times 6}$ and measure their distance error as $e'_{D}(\mathbf{v})=\|x_\text{obs}-x_\text{recon}\|_2$ and their orientation error as $e'_{N}(\mathbf{v})=\text{acos}(n_\text{obs}^Tn_\text{recon})$. This gives a combined coverage score of:
\begin{equation}
    \vspace{\vspacebeforeequation}
    S_2 = \frac{1}{|\mathcal{V}|} \left|\left\{
    \mathbf{v}\in \mathcal{V} | e'_{D}(\mathbf{v})<\delta_D, e'_{N}(\mathbf{v}) < \delta_N
    \right\}\right|.
    \label{eq:coverage_score}
    \vspace{\vspaceafterequation}
\end{equation}
We use $S_1$ as the main confidence measurement and $S_2$ for avoiding poorly observed cases, so the final confidence score $C$ is:
\begin{equation}
    \vspace{\vspacebeforeequation}
    C = S_1 * \max(1.0, S_2 / \delta_C),
    \label{eq:confidence}
    \vspace{\vspaceafterequation}
\end{equation}
where $\delta_C\in [0.0,1.0]$ is a threshold controlling the importance of the coverage score.
Note how the above confidence values evaluate the quality of the output shown on the right in Fig.~\ref{fig:method}, which enables the user to select the balance between recall and precision during inference.

Finally, for each point in the scene point cloud $X_{N\times 6}$, we check whether the distance error in Eq.~\ref{eq:error_dist} and Eq.~\ref{eq:error_normal} is smaller than the output threshold
and mark points with small errors as in the foreground. Note that since the observation is noisy and the learned shape prior is not perfect, the output error thresholds can be larger than the ones used in Eq.~\ref{eq:fitting_score} and Eq.\ref{eq:coverage_score}.

\subsection{Multiple Proposals}
\label{sec:full_alg}
Since the EM algorithm outputs can be affected by its initialization and we do not know the number of objects in the scene,
we initialize a large number of proposals randomly spread across the entire scene to cover all possible objects.
We observe that many proposals will quickly converge to similar positions during the early iterations. Therefore, at each iteration, we remove duplicated proposals and only keep the one with the highest fitting score $S_1$ defined in Eq.~\ref{eq:fitting_score}.
Duplication is determined by computing the overlap of $W$ between different proposals (see Supp. for details). 

As shown in Fig.~\ref{fig:method}, we further divide the iterations into two phases. 
In the first phase, since the early shape estimation is not converged to a reasonable place, we let each proposal run fully independently.
When we enter the second phase, mesh extraction and pose estimation will be first applied to remove proposals with unreasonable scales. Then, we optimize all proposals globally by updating the joint assignment weight with:
\begin{equation}
    \vspace{\vspacebeforeequation}
    \vspace{\vspacebeforeequation}
    W^{(k)}_t[i] = \frac{S_1^{(k)}e^{-E(X[i], \latent^{(k)}_t)}}{\sum_{j}S_1^{(j)}e^{-E(X[i], \latent^{(j)}_t)} + \Omega},\label{eq:w_update_joint}
    \vspace{\vspaceafterequation}
    \vspace{\vspaceafterequation}
\end{equation}
where $k$ is the index of current active proposals and $S_1^{(k)}$ is the current fitting score in Eq.~\ref{eq:fitting_score}, which increases the assignment weight for more confident proposals.
During the last iterations in Phase-2, we also remove the proposals that are largely contained by other proposals to simplify the decomposition of the scene, following a similar methodology to duplication removal. 
Additional details of this process are available in the supplemental.

\vspace{\vspacebeforesection}
\section{Experiments}
\vspace{\vspaceaftersection}

\begin{figure*}[t!]
\centering
    \vspace{-.2em}
   \includegraphics[width=1.0\textwidth]{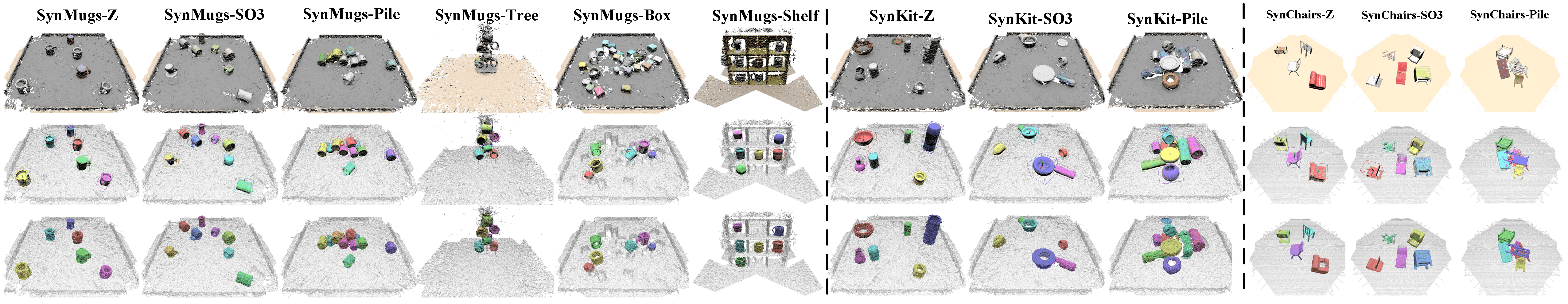}
   \\\vspace{-.7em}
    \caption{\textbf{Synthetic scenes:} \textbf{Top}: the mesh reconstruction as input, note how our data has a realistic simulated depth pattern; \textbf{Middle}: visualization of the estimated shapes, poses and bounding boxes as the byproducts of our method. \textbf{Bottom}: Our segmentation prediction.}
    \label{fig:syn_results}
\end{figure*}

\begin{figure*}[t!]
\centering
    \vspace{-.2em}
   \includegraphics[width=1.0\textwidth]{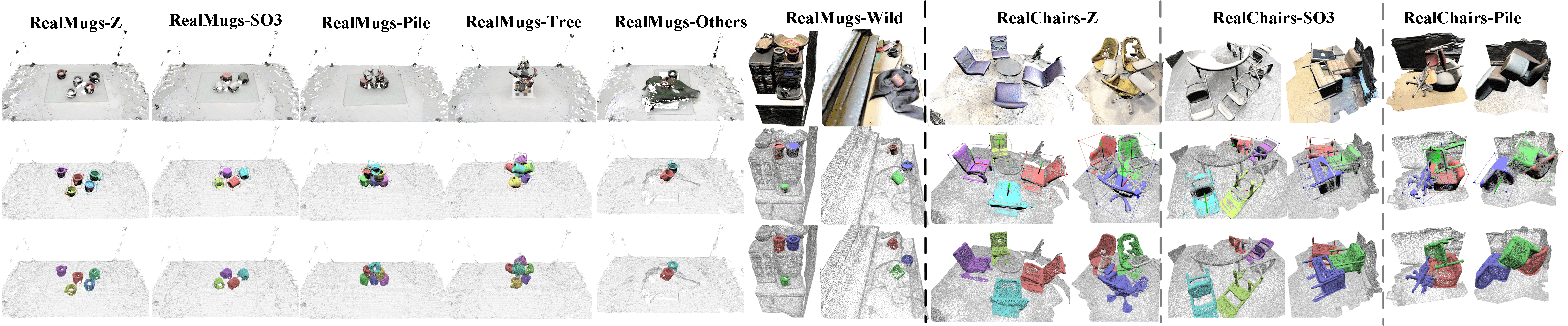}
   \\\vspace{-.7em}
    \caption{\textbf{Chairs and Mugs} real testset: the three rows are in the same format as Fig.~\ref{fig:syn_results}.
    }
    \label{fig:real_results}
\end{figure*}

We focus our experiments on answering four main questions: First, can our method successfully segment the objects of interest from the scene?
Second, how does our method compare to existing baselines when different training data distributions are accessible?
Third, how does each approach generalize to the different testing environments, including the real-world scenes and the out-of-distribution configurations?
And fourth, how do the different components of our model contribute to its performance?
To answer these questions, we perform a series of baseline comparisons as well as ablation studies both on synthetic and real-world scenes.

\vspace{\vspacebeforesection}
\subsection{Baselines}
\vspace{\vspaceaftersection}
To the best of the authors' knowledge, fully unsupervised instance segmentation (not foreground-background segmentation) in static scenes remains unexplored in the deep learning literature. Therefore, we compare our method with supervised and weakly supervised methods. 
SoftGroup (CVPR22)~\cite{vu2022softgroup} is the current SoTA for 3D instance segmentation methods that are trained with ground truth instance masks.
Box2Mask (ECCV22)~\cite{chibane2022box2mask} is a recent weakly supervised method, which only needs ground truth instance bounding boxes as supervision while retaining competitive performance.
The closest weakly supervised method to ours is ContrastiveSceneContext (CVPR21)~\cite{hou2021exploring}, which can be trained with only a few point labels.
To conduct fair comparisons and reduce the sim2real gap, all baselines are fully trained on the corresponding training set with positions and normals as input while not using colors.
\begin{table*}[t]
\centering
\scalebox{0.6}{
\begin{tabular}{ll|cc|cc|cc|cc|cc|cc|cc|cc|cc|cc|cc|cc}
\toprule
\textbf{Mugs} & \textbf{Testing} & \multicolumn{2}{c|}{\textbf{Z}} & \multicolumn{2}{c|}{\textbf{SO3}} & \multicolumn{2}{c|}{\textbf{Pile}} & \multicolumn{2}{c|}{\textbf{Tree}} & \multicolumn{2}{c|}{\textbf{Box}} & \multicolumn{2}{c|}{\textbf{Shelf}} & \multicolumn{2}{c|}{\textbf{(R) Z}} & \multicolumn{2}{c|}{\textbf{(R) SO3}} & \multicolumn{2}{c|}{\cellcolor[HTML]{FFFFFF}\textbf{(R) Pile}} & \multicolumn{2}{c|}{\cellcolor[HTML]{FFFFFF}\textbf{(R) Tree}} & \multicolumn{2}{c|}{\textbf{(R) Others}} & \multicolumn{2}{c}{\textbf{(R) Wild}} \\ \midrule
\textbf{Training} & Metrics & AP & A50 & AP & A50 & AP & A50 & AP & A50 & AP & A50 & AP & A50 & AP & A50 & AP & A50 & AP & A50 & AP & A50 & AP & A50 & AP & A50 \\ \midrule
 & CSC (100) & 6.0 & 22.0 & \cellcolor[HTML]{D9D9D9}1.9 & \cellcolor[HTML]{D9D9D9}7.6 & \cellcolor[HTML]{D9D9D9}0.1 & \cellcolor[HTML]{D9D9D9}0.5 & \cellcolor[HTML]{D9D9D9}0.0 & \cellcolor[HTML]{D9D9D9}0.0 & \cellcolor[HTML]{D9D9D9}0.0 & \cellcolor[HTML]{D9D9D9}0.2 & \cellcolor[HTML]{D9D9D9}0.0 & \cellcolor[HTML]{D9D9D9}0.0 & 0.0 & 0.0 & \cellcolor[HTML]{D9D9D9}0.0 & \cellcolor[HTML]{D9D9D9}0.0 & \cellcolor[HTML]{D9D9D9}0.0 & \cellcolor[HTML]{D9D9D9}0.0 & \cellcolor[HTML]{D9D9D9}0.0 & \cellcolor[HTML]{D9D9D9}0.0 & \cellcolor[HTML]{D9D9D9}0.0 & \cellcolor[HTML]{D9D9D9}0.0 & \cellcolor[HTML]{D9D9D9}0.2 & \cellcolor[HTML]{D9D9D9}0.9 \\
 & CSC (200) & 78.7 & 98.1 & \cellcolor[HTML]{D9D9D9}62.7 & \cellcolor[HTML]{D9D9D9}83.3 & \cellcolor[HTML]{D9D9D9}8.9 & \cellcolor[HTML]{D9D9D9}17.4 & \cellcolor[HTML]{D9D9D9}0.6 & \cellcolor[HTML]{D9D9D9}2.0 & \cellcolor[HTML]{D9D9D9}2.0 & \cellcolor[HTML]{D9D9D9}5.0 & \cellcolor[HTML]{D9D9D9}0.8 & \cellcolor[HTML]{D9D9D9}3.1 & 72.1 & 99.8 & \cellcolor[HTML]{D9D9D9}9.5 & \cellcolor[HTML]{D9D9D9}21.4 & \cellcolor[HTML]{D9D9D9}0.3 & \cellcolor[HTML]{D9D9D9}1.6 & \cellcolor[HTML]{D9D9D9}0.6 & \cellcolor[HTML]{D9D9D9}2.8 & \cellcolor[HTML]{D9D9D9}6.7 & \cellcolor[HTML]{D9D9D9}16.0 & \cellcolor[HTML]{D9D9D9}2.0 & \cellcolor[HTML]{D9D9D9}6.1 \\
 & Box2Mask & 96.9 & \textbf{100} & \cellcolor[HTML]{D9D9D9}92.1 & \cellcolor[HTML]{D9D9D9}99.3 & \cellcolor[HTML]{D9D9D9}36.7 & \cellcolor[HTML]{D9D9D9}27.6 & \cellcolor[HTML]{D9D9D9}12.3 & \cellcolor[HTML]{D9D9D9}45.3 & \cellcolor[HTML]{D9D9D9}10.0 & \cellcolor[HTML]{D9D9D9}24.0 & \cellcolor[HTML]{D9D9D9}8.6 & \cellcolor[HTML]{D9D9D9}14.6 & 98.1 & \textbf{100} & \cellcolor[HTML]{D9D9D9}93.1 & \cellcolor[HTML]{D9D9D9}{\ul \textbf{100}} & \cellcolor[HTML]{D9D9D9}5.6 & \cellcolor[HTML]{D9D9D9}19.1 & \cellcolor[HTML]{D9D9D9}8.4 & \cellcolor[HTML]{D9D9D9}28.8 & \cellcolor[HTML]{D9D9D9}18.4 & \cellcolor[HTML]{D9D9D9}39.5 & \cellcolor[HTML]{D9D9D9}18.1 & \cellcolor[HTML]{D9D9D9}27.9 \\
\multirow{-4}{*}{\textbf{Scene Z}} & SoftGroup & \textbf{100} & \textbf{100} & \cellcolor[HTML]{D9D9D9}{\ul 96.5} & \cellcolor[HTML]{D9D9D9}98.5 & \cellcolor[HTML]{D9D9D9}21.9 & \cellcolor[HTML]{D9D9D9}27.6 & \cellcolor[HTML]{D9D9D9}0.4 & \cellcolor[HTML]{D9D9D9}1.8 & \cellcolor[HTML]{D9D9D9}1.6 & \cellcolor[HTML]{D9D9D9}3.7 & \cellcolor[HTML]{D9D9D9}9.2 & \cellcolor[HTML]{D9D9D9}17.2 & 93.7 & 97.6 & \cellcolor[HTML]{D9D9D9}{\ul 97.3} & \cellcolor[HTML]{D9D9D9}{\ul \textbf{100}} & \cellcolor[HTML]{D9D9D9}0.0 & \cellcolor[HTML]{D9D9D9}0.0 & \cellcolor[HTML]{D9D9D9}0.9 & \cellcolor[HTML]{D9D9D9}3.4 & \cellcolor[HTML]{D9D9D9}4.3 & \cellcolor[HTML]{D9D9D9}9.1 & \cellcolor[HTML]{D9D9D9}18.3 & \cellcolor[HTML]{D9D9D9}32.9 \\ \midrule
 & CSC (100) & 1.7 & 8.6 & 1.4 & 6.5 & \cellcolor[HTML]{D9D9D9}0.1 & \cellcolor[HTML]{D9D9D9}0.6 & \cellcolor[HTML]{D9D9D9}0.0 & \cellcolor[HTML]{D9D9D9}0.0 & \cellcolor[HTML]{D9D9D9}0.0 & \cellcolor[HTML]{D9D9D9}0.1 & \cellcolor[HTML]{D9D9D9}0.1 & \cellcolor[HTML]{D9D9D9}0.6 & 0.0 & 0.0 & 0.0 & 0.0 & \cellcolor[HTML]{D9D9D9}0.0 & \cellcolor[HTML]{D9D9D9}0.0 & \cellcolor[HTML]{D9D9D9}0.0 & \cellcolor[HTML]{D9D9D9}0.0 & \cellcolor[HTML]{D9D9D9}0.0 & \cellcolor[HTML]{D9D9D9}0.0 & \cellcolor[HTML]{D9D9D9}0.7 & \cellcolor[HTML]{D9D9D9}1.9 \\
 & CSC (200) & 70.0 & 99.3 & 72.7 & 95.7 & \cellcolor[HTML]{D9D9D9}22.3 & \cellcolor[HTML]{D9D9D9}42.2 & \cellcolor[HTML]{D9D9D9}10.2 & \cellcolor[HTML]{D9D9D9}24.6 & \cellcolor[HTML]{D9D9D9}5.4 & \cellcolor[HTML]{D9D9D9}12.3 & \cellcolor[HTML]{D9D9D9}5.2 & \cellcolor[HTML]{D9D9D9}20.0 & 68.4 & \textbf{100} & 59.8 & \textbf{100} & \cellcolor[HTML]{D9D9D9}1.0 & \cellcolor[HTML]{D9D9D9}3.7 & \cellcolor[HTML]{D9D9D9}4.8 & \cellcolor[HTML]{D9D9D9}20.1 & \cellcolor[HTML]{D9D9D9}6.9 & \cellcolor[HTML]{D9D9D9}19.7 & \cellcolor[HTML]{D9D9D9}3.9 & \cellcolor[HTML]{D9D9D9}11.8 \\
 & Box2Mask & 95.8 & \textbf{100} & 94.7 & \textbf{100} & \cellcolor[HTML]{D9D9D9}53.7 & \cellcolor[HTML]{D9D9D9}80.1 & \cellcolor[HTML]{D9D9D9}30.2 & \cellcolor[HTML]{D9D9D9}80.6 & \cellcolor[HTML]{D9D9D9}26.5 & \cellcolor[HTML]{D9D9D9}40.1 & \cellcolor[HTML]{D9D9D9}13.4 & \cellcolor[HTML]{D9D9D9}33.0 & \textbf{100} & \textbf{100} & 95.3 & \textbf{100} & \cellcolor[HTML]{D9D9D9}12.7 & \cellcolor[HTML]{D9D9D9}34.3 & \cellcolor[HTML]{D9D9D9}19.6 & \cellcolor[HTML]{D9D9D9}49.2 & \cellcolor[HTML]{D9D9D9}27.1 & \cellcolor[HTML]{D9D9D9}50.6 & \cellcolor[HTML]{D9D9D9}50.3 & \cellcolor[HTML]{D9D9D9}74.8 \\
\multirow{-4}{*}{\textbf{Scene SO3}} & SoftGroup & \textbf{100} & \textbf{100} & 99.6 & 99.8 & \cellcolor[HTML]{D9D9D9}43.9 & \cellcolor[HTML]{D9D9D9}51.9 & \cellcolor[HTML]{D9D9D9}10.0 & \cellcolor[HTML]{D9D9D9}18.7 & \cellcolor[HTML]{D9D9D9}15.3 & \cellcolor[HTML]{D9D9D9}21.3 & \cellcolor[HTML]{D9D9D9}20.8 & \cellcolor[HTML]{D9D9D9}30.9 & 99.2 & \textbf{100} & 98.2 & \textbf{100} & \cellcolor[HTML]{D9D9D9}2.8 & \cellcolor[HTML]{D9D9D9}4.8 & \cellcolor[HTML]{D9D9D9}10.4 & \cellcolor[HTML]{D9D9D9}21.7 & \cellcolor[HTML]{D9D9D9}6.5 & \cellcolor[HTML]{D9D9D9}13.3 & \cellcolor[HTML]{D9D9D9}18.3 & \cellcolor[HTML]{D9D9D9}30.7 \\ \midrule
 & CSC (100) & 0.0 & 0.3 & 0.0 & 0.3 & 0.0 & 0.1 & \cellcolor[HTML]{D9D9D9}0.0 & \cellcolor[HTML]{D9D9D9}0.0 & \cellcolor[HTML]{D9D9D9}0.0 & \cellcolor[HTML]{D9D9D9}0.0 & \cellcolor[HTML]{D9D9D9}0.0 & \cellcolor[HTML]{D9D9D9}0.0 & 0.0 & 0.0 & \multicolumn{1}{r}{0.0} & \multicolumn{1}{r|}{0.0} & 0.0 & 0.0 & \cellcolor[HTML]{D9D9D9}0.0 & \cellcolor[HTML]{D9D9D9}0.0 & \cellcolor[HTML]{D9D9D9}0.0 & \cellcolor[HTML]{D9D9D9}0.0 & \cellcolor[HTML]{D9D9D9}0.7 & \cellcolor[HTML]{D9D9D9}1.8 \\
 & CSC (200) & 53.3 & 88.4 & 50.2 & 77.7 & 29.5 & 61.3 & \cellcolor[HTML]{D9D9D9}29.4 & \cellcolor[HTML]{D9D9D9}66.2 & \cellcolor[HTML]{D9D9D9}7.6 & \cellcolor[HTML]{D9D9D9}19.8 & \cellcolor[HTML]{D9D9D9}10.3 & \cellcolor[HTML]{D9D9D9}34.1 & 13.0 & 45.0 & 22.7 & 65.2 & 5.3 & 30.1 & \cellcolor[HTML]{D9D9D9}1.5 & \cellcolor[HTML]{D9D9D9}8.0 & \cellcolor[HTML]{D9D9D9}3.5 & \cellcolor[HTML]{D9D9D9}15.4 & \cellcolor[HTML]{D9D9D9}2.8 & \cellcolor[HTML]{D9D9D9}8.5 \\
 & Box2Mask & 96.0 & \textbf{100} & 94.7 & \textbf{100} & 78.7 & \textbf{99.6} & \cellcolor[HTML]{D9D9D9}55.2 & \cellcolor[HTML]{D9D9D9}94.8 & \cellcolor[HTML]{D9D9D9}42.7 & \cellcolor[HTML]{D9D9D9}52.9 & \cellcolor[HTML]{D9D9D9}26.3 & \cellcolor[HTML]{D9D9D9}54.6 & \textbf{100} & \textbf{100} & 95.8 & \textbf{100} & 72.3 & \textbf{98.0} & \cellcolor[HTML]{D9D9D9}55.8 & \cellcolor[HTML]{D9D9D9}\textbf{98.1} & \cellcolor[HTML]{D9D9D9}56.0 & \cellcolor[HTML]{D9D9D9}79.2 & \cellcolor[HTML]{D9D9D9}45.1 & \cellcolor[HTML]{D9D9D9}71.4 \\
\multirow{-4}{*}{\textbf{Scene Pile}} & SoftGroup & 99.5 & \textbf{100} & \textbf{99.7} & \textbf{100} & \textbf{89.0} & 93.2 & \cellcolor[HTML]{D9D9D9}42.3 & \cellcolor[HTML]{D9D9D9}72.8 & \cellcolor[HTML]{D9D9D9}23.4 & \cellcolor[HTML]{D9D9D9}25.5 & \cellcolor[HTML]{D9D9D9}28.8 & \cellcolor[HTML]{D9D9D9}39.5 & 99.6 & \textbf{100} & \textbf{99.7} & \textbf{100} & \textbf{74.7} & 86.3 & \cellcolor[HTML]{D9D9D9}53.4 & \cellcolor[HTML]{D9D9D9}83.0 & \cellcolor[HTML]{D9D9D9}50.9 & \cellcolor[HTML]{D9D9D9}66.5 & \cellcolor[HTML]{D9D9D9}48.1 & \cellcolor[HTML]{D9D9D9}65.9 \\ \midrule
\textbf{ShapeNet} & EFEM & \cellcolor[HTML]{D9D9D9}78.4 & \cellcolor[HTML]{D9D9D9}99.8 & \cellcolor[HTML]{D9D9D9}79.3 & \cellcolor[HTML]{D9D9D9}{\ul 99.8} & \cellcolor[HTML]{D9D9D9}{\ul 68.2} & \cellcolor[HTML]{D9D9D9}{\ul 96.8} & \cellcolor[HTML]{D9D9D9}\textbf{68.8} & \cellcolor[HTML]{D9D9D9}\textbf{99.0} & \cellcolor[HTML]{D9D9D9}\textbf{59.9} & \cellcolor[HTML]{D9D9D9}\textbf{77.0} & \cellcolor[HTML]{D9D9D9}\textbf{48.7} & \cellcolor[HTML]{D9D9D9}\textbf{72.4} & \cellcolor[HTML]{D9D9D9}87.2 & \cellcolor[HTML]{D9D9D9}\textbf{100} & \cellcolor[HTML]{D9D9D9}87.3 & \cellcolor[HTML]{D9D9D9}{\ul \textbf{100}} & \cellcolor[HTML]{D9D9D9}{\ul 63.1} & \cellcolor[HTML]{D9D9D9}{\ul 94.0} & \cellcolor[HTML]{D9D9D9}\textbf{69.6} & \cellcolor[HTML]{D9D9D9}95.4 & \cellcolor[HTML]{D9D9D9}\textbf{69.7} & \cellcolor[HTML]{D9D9D9}\textbf{89.4} & \cellcolor[HTML]{D9D9D9}\textbf{54.1} & \cellcolor[HTML]{D9D9D9}\textbf{82.3} \\
\bottomrule
\end{tabular}
}
\vspace{-.7em}
\caption{
Results ($\%$) on SynMugs (left) and RealMugs (right with label R). A50 corresponds to mAP50. The full table including mAP25 is available in the supplemental. 
The grey cells highlight that the testing scenes are out of the training distribution and the bold number means the best among all methods while the underlined number means the best within grey cells.}
\label{tab:compact_mugs}
\end{table*}
\begin{table}[t]
\centering
\scalebox{0.5}{
\begin{tabular}{ll|cc|cc|cc|cc|cc|cc}
\toprule
\multicolumn{1}{l}{\textbf{}} & \textbf{Testing} & \multicolumn{2}{c|}{\textbf{(C) Z}} & \multicolumn{2}{c|}{\textbf{(C) SO3}} & \multicolumn{2}{c|}{\textbf{(C) Pile}} & \multicolumn{2}{c|}{\textbf{(K) Z}} & \multicolumn{2}{c|}{\textbf{(K) SO3}} & \multicolumn{2}{c}{\textbf{(K) Pile}} \\ \midrule
\textbf{Training} & Metrics & AP & A50 & AP & A50 & AP & A50 & AP & A50 & AP & A50 & AP & A50 \\ \midrule
 & CSC (100) & 66.2 & 83.1 & \cellcolor[HTML]{D9D9D9}56.6 & \cellcolor[HTML]{D9D9D9}80.2 & \cellcolor[HTML]{D9D9D9}6.1 & \cellcolor[HTML]{D9D9D9}15.2 & 12.7 & 29.9 & \cellcolor[HTML]{D9D9D9}7.5 & \cellcolor[HTML]{D9D9D9}14.5 & \cellcolor[HTML]{D9D9D9}2.3 & \cellcolor[HTML]{D9D9D9}5.9 \\
 & CSC (200) & 77.7 & 91.0 & \cellcolor[HTML]{D9D9D9}63.7 & \cellcolor[HTML]{D9D9D9}85.1 & \cellcolor[HTML]{D9D9D9}10.5 & \cellcolor[HTML]{D9D9D9}22.7 & 63.0 & 80.9 & \cellcolor[HTML]{D9D9D9}37.9 & \cellcolor[HTML]{D9D9D9}49.2 & \cellcolor[HTML]{D9D9D9}12.2 & \cellcolor[HTML]{D9D9D9}23.2 \\
 & Box2Mask & \textbf{99.9} & \textbf{100} & \cellcolor[HTML]{D9D9D9}{\ul 95.8} & \cellcolor[HTML]{D9D9D9}{\ul 98.9} & \cellcolor[HTML]{D9D9D9}19.3 & \cellcolor[HTML]{D9D9D9}41.2 & 89.7 & 97.3 & \cellcolor[HTML]{D9D9D9}69.6 & \cellcolor[HTML]{D9D9D9}87.0 & \cellcolor[HTML]{D9D9D9}41.4 & \cellcolor[HTML]{D9D9D9}68.9 \\
\multirow{-4}{*}{\textbf{Scene Z}} & SoftGroup & 99.8 & \textbf{100} & \cellcolor[HTML]{D9D9D9}94.8 & \cellcolor[HTML]{D9D9D9}98.5 & \cellcolor[HTML]{D9D9D9}24.1 & \cellcolor[HTML]{D9D9D9}36.5 & 93.6 & 96.8 & \cellcolor[HTML]{D9D9D9}{\ul 94.0} & \cellcolor[HTML]{D9D9D9}{\ul 99.3} & \cellcolor[HTML]{D9D9D9}52.3 & \cellcolor[HTML]{D9D9D9}63.5 \\ \midrule
 & CSC (100) & 74.9 & 81.4 & 77.3 & 82.4 & \cellcolor[HTML]{D9D9D9}17.7 & \cellcolor[HTML]{D9D9D9}31.0 & 8.5 & 21.7 & 11.3 & 23.8 & \cellcolor[HTML]{D9D9D9}2.0 & \cellcolor[HTML]{D9D9D9}5.4 \\
 & CSC (200) & 84.3 & 86.6 & 85.4 & 88.5 & \cellcolor[HTML]{D9D9D9}20.6 & \cellcolor[HTML]{D9D9D9}35.3 & 27.8 & 48.9 & 53.9 & 77.5 & \cellcolor[HTML]{D9D9D9}18.1 & \cellcolor[HTML]{D9D9D9}38.9 \\
 & Box2Mask & 99.7 & \textbf{100} & 99.1 & 99.5 & \cellcolor[HTML]{D9D9D9}50.5 & \cellcolor[HTML]{D9D9D9}80.3 & 85.3 & 96.2 & 91.8 & 98.5 & \cellcolor[HTML]{D9D9D9}52.9 & \cellcolor[HTML]{D9D9D9}76.5 \\
\multirow{-4}{*}{\textbf{Scene SO3}} & SoftGroup & 99.6 & \textbf{100} & 99.1 & \textbf{100} & \cellcolor[HTML]{D9D9D9}55.1 & \cellcolor[HTML]{D9D9D9}68.2 & 89.3 & 93.9 & \textbf{96.9} & \textbf{99.3} & \cellcolor[HTML]{D9D9D9}56.2 & \cellcolor[HTML]{D9D9D9}67.2 \\ \midrule
 & CSC (100) & 67.0 & 74.1 & 74.6 & 82.1 & 47.2 & 63.8 & 3.4 & 11.2 & 2.6 & 7.1 & 1.4 & 4.5 \\
 & CSC (200) & 71.7 & 76.3 & 88.4 & 92.4 & 67.4 & 81.6 & 26.5 & 53.8 & 44.8 & 73.6 & 28.8 & 60.4 \\
 & Box2Mask & 99.6 & 99.7 & \textbf{99.5} & 99.7 & 78.6 & 96.9 & 87.2 & \textbf{97.9} & 92.1 & 98.6 & 74.8 & \textbf{94.6} \\
\multirow{-4}{*}{\textbf{Scene Pile}} & SoftGroup & 99.4 & \textbf{100} & 98.9 & \textbf{100} & \textbf{95.0} & \textbf{97.0} & \textbf{93.7} & 97.3 & 96.0 & 98.8 & \textbf{84.6} & 91.3 \\ \midrule
\textbf{ShapeNet} & EFEM & \cellcolor[HTML]{D9D9D9}93.1 & \cellcolor[HTML]{D9D9D9}99.2 & \cellcolor[HTML]{D9D9D9}86.1 & \cellcolor[HTML]{D9D9D9}97.4 & \cellcolor[HTML]{D9D9D9}{\ul 75.3} & \cellcolor[HTML]{D9D9D9}{\ul 88.0} & \cellcolor[HTML]{D9D9D9}69.4 & \cellcolor[HTML]{D9D9D9}83.4 & \cellcolor[HTML]{D9D9D9}67.6 & \cellcolor[HTML]{D9D9D9}83.1 & \cellcolor[HTML]{D9D9D9}{\ul 60.1} & \cellcolor[HTML]{D9D9D9}{\ul 78.9} \\
\bottomrule
\end{tabular}
}
\vspace{-.7em}
\caption{ 
SynChairs (left) and SynKit (right), same format as Tab.~\ref{tab:compact_mugs}}
\label{tab:compact_syn_chair_kit}
\end{table}

\begin{table}[t]
\centering
\scalebox{0.55}{
\begin{tabular}{ll|ccc|ccc|ccc}
\toprule
\textbf{RealChairs} & \textbf{Testing} & \multicolumn{3}{c|}{\textbf{R(Z)}} & \multicolumn{3}{c|}{\textbf{R(SO3)}} & \multicolumn{3}{c}{\textbf{R(Pile)}} \\ \midrule
\textbf{Training} & Metrics & AP & AP50 & AP25 & AP & AP50 & AP25 & AP & AP50 & AP25 \\ \midrule
 & CSC (100) & 0.5 & 1.4 & 9.4 & \cellcolor[HTML]{D9D9D9}1.3 & \cellcolor[HTML]{D9D9D9}2.8 & \cellcolor[HTML]{D9D9D9}10.8 & \cellcolor[HTML]{D9D9D9}0.0 & \cellcolor[HTML]{D9D9D9}0.0 & \cellcolor[HTML]{D9D9D9}7.2 \\
 & CSC (200) & 0.8 & 1.2 & 11.0 & \cellcolor[HTML]{D9D9D9}0.3 & \cellcolor[HTML]{D9D9D9}0.6 & \cellcolor[HTML]{D9D9D9}9.8 & \cellcolor[HTML]{D9D9D9}0.0 & \cellcolor[HTML]{D9D9D9}0.2 & \cellcolor[HTML]{D9D9D9}3.1 \\
 & Box2Mask & 0.0 & 0.3 & 20.2 & \cellcolor[HTML]{D9D9D9}0.3 & \cellcolor[HTML]{D9D9D9}1.9 & \cellcolor[HTML]{D9D9D9}27.2 & \cellcolor[HTML]{D9D9D9}0.0 & \cellcolor[HTML]{D9D9D9}0.1 & \cellcolor[HTML]{D9D9D9}11.8 \\
\multirow{-4}{*}{\textbf{Scene Z}} & SoftGroup & 5.1 & 7.7 & 27.0 & \cellcolor[HTML]{D9D9D9}2.8 & \cellcolor[HTML]{D9D9D9}6.7 & \cellcolor[HTML]{D9D9D9}21.9 & \cellcolor[HTML]{D9D9D9}0.0 & \cellcolor[HTML]{D9D9D9}0.0 & \cellcolor[HTML]{D9D9D9}3.7 \\ \midrule
 & CSC (100) & 3.6 & 5.3 & 18.2 & 1.4 & 2.5 & 13.3 & \cellcolor[HTML]{D9D9D9}0.0 & \cellcolor[HTML]{D9D9D9}0.1 & \cellcolor[HTML]{D9D9D9}4.5 \\
 & CSC (200) & 1.2 & 2.6 & 10.4 & 1.8 & 4.9 & 20.2 & \cellcolor[HTML]{D9D9D9}0.1 & \cellcolor[HTML]{D9D9D9}0.5 & \cellcolor[HTML]{D9D9D9}9.1 \\
 & Box2Mask & 1.5 & 5.9 & 48.4 & 5.1 & 19.9 & 56.7 & \cellcolor[HTML]{D9D9D9}0.4 & \cellcolor[HTML]{D9D9D9}1.8 & \cellcolor[HTML]{D9D9D9}28.2 \\
\multirow{-4}{*}{\textbf{Scene SO3}} & SoftGroup & 13.8 & 17.2 & 27.4 & 7.9 & 16.2 & 30.3 & \cellcolor[HTML]{D9D9D9}0.1 & \cellcolor[HTML]{D9D9D9}0.1 & \cellcolor[HTML]{D9D9D9}6.1 \\ \midrule
 & CSC (100) & 5.2 & 7.1 & 9.6 & 3.2 & 7.4 & 12.3 & 0.1 & 0.3 & 2.9 \\
 & CSC (200) & 8.3 & 11.7 & 23.1 & 13.9 & 23.2 & 33.6 & 1.0 & 2.6 & 13.1 \\
 & Box2Mask & 1.6 & 6.5 & 37.8 & 9.4 & 22.0 & 65.3 & 1.3 & 4.8 & 46.6 \\
\multirow{-4}{*}{\textbf{Scene Pile}} & SoftGroup & 20.6 & 27.8 & 36.1 & 14.4 & 24.4 & 42.1 & 0.4 & 1.5 & 8.0 \\ \midrule
 & CSC (100) & 36.8 & 49.8 & 62.0 & \cellcolor[HTML]{D9D9D9}5.3 & \cellcolor[HTML]{D9D9D9}13.1 & \cellcolor[HTML]{D9D9D9}18.8 & \cellcolor[HTML]{D9D9D9}3.9 & \cellcolor[HTML]{D9D9D9}10.8 & \cellcolor[HTML]{D9D9D9}18.1 \\
 & CSC (200) & 50.0 & 74.3 & 77.8 & \cellcolor[HTML]{D9D9D9}6.2 & \cellcolor[HTML]{D9D9D9}13.1 & \cellcolor[HTML]{D9D9D9}18.6 & \cellcolor[HTML]{D9D9D9}5.5 & \cellcolor[HTML]{D9D9D9}13.4 & \cellcolor[HTML]{D9D9D9}27.1 \\
 & Box2Mask & \textbf{84.1} & \textbf{99.4} & \textbf{99.4} & \cellcolor[HTML]{D9D9D9}16.2 & \cellcolor[HTML]{D9D9D9}33.7 & \cellcolor[HTML]{D9D9D9}41.2 & \cellcolor[HTML]{D9D9D9}14.3 & \cellcolor[HTML]{D9D9D9}29.8 & \cellcolor[HTML]{D9D9D9}50.7 \\
\multirow{-4}{*}{\textbf{Scannet}} & SoftGroup & 74.0 & 81.1 & 87.5 & \cellcolor[HTML]{D9D9D9}22.3 & \cellcolor[HTML]{D9D9D9}39.6 & \cellcolor[HTML]{D9D9D9}48.6 & \cellcolor[HTML]{D9D9D9}11.5 & \cellcolor[HTML]{D9D9D9}18.0 & \cellcolor[HTML]{D9D9D9}36.1 \\ \midrule
\textbf{ShapeNet} & EFEM & \cellcolor[HTML]{D9D9D9}51.0 & \cellcolor[HTML]{D9D9D9}78.3 & \cellcolor[HTML]{D9D9D9}85.0 & \cellcolor[HTML]{D9D9D9}{\ul \textbf{51.2}} & \cellcolor[HTML]{D9D9D9}{\ul \textbf{75.7}} & \cellcolor[HTML]{D9D9D9}{\ul \textbf{87.9}} & \cellcolor[HTML]{D9D9D9}{\ul \textbf{34.4}} & \cellcolor[HTML]{D9D9D9}{\ul \textbf{55.3}} & \cellcolor[HTML]{D9D9D9}{\ul \textbf{74.8}} \\ \bottomrule
\end{tabular}
}
\vspace{-.7em}
\caption{
Results on RealChairs, same format as Tab.~\ref{tab:compact_mugs}} 
\label{tab:compact_real_chairs}
\end{table}

\vspace{\vspacebeforesection}
\subsection{Experimental Setup}
\label{sec:exp_setup}
\vspace{\vspaceaftersection}
We focus our experiments on scenes that contain  objects humans or robots can interact with.  These objects are more interesting targets for segmentation as their configurations can change dramatically in real-world applications, such as robotics or AR/VR.
We experiment with three object categories that frequently appear in the presence of clutter and in diverse configurations in real-world data: \textbf{Mugs}, \textbf{Kitchen containers}, and \textbf{Chairs}.
Since there is no available dataset that contains these objects with rich configuration changes (not just sitting upright on a flat surface), and all baselines need to be trained with ground truth, we gather simulated data to train the baselines, then evaluate on simulated and real test scenes containing completely unseen object instances.

Three types of scenes are used for training baselines on each of the object classes. Take the mugs scene as an example (Fig.~\ref{fig:syn_results} Left top): 
In the \textbf{Z} scenes, all instances are upright and not in contact with each other.
In \textbf{SO(3)}, the objects can have a random orientation but are still not in contact with each other. 
\textbf{Pile} is a much more challenging setting where the objects can touch each other, can take any orientation, and can form piles.
We simulate 500 scenes for training, 50 scenes for validation and 100 scenes for testing in each of the setups.
For each baseline we train three models, one on the data from each of the three scene types.

We train our equivariant shape prior (Sec.~\ref{sec:shape_prior}) on the corresponding categories from ShapeNet~\cite{chang2015shapenet} and freeze their weights once trained. Note that all the following results are generated from the same trained shape prior for each category, and all the objects in our testing scenes never appear in the training set of the shape prior. 
We evaluate each model on all of the scenes and report results in Tab.~\ref{tab:compact_mugs}, Tab.~\ref{tab:compact_syn_chair_kit} and Tab.~\ref{tab:compact_real_chairs}.
The gray cells indicate results from models that were trained on less difficult datasets than they were evaluated on, with the underlined number indicating the best performing inside gray cells. The metric  for evaluating the segmentation is the commonly used mAP~\cite{dai2017scannet}.

\vspace{\vspacebeforesection}
\subsection{Results on synthetic data}
\label{sec:exp_syn}
\vspace{\vspaceaftersection}

We first evaluate the performance on simulated data, and we found that existing baselines work very well inside the training distribution if trained with enough supervisory signals. 
However, the weakly supervised method CSC~\cite{hou2021exploring} has a significant performance drop when the number of supervision points decreases from 200 to 100.
This performance drop becomes increasingly severe when the training set distribution becomes more complex, demonstrating the difficulty of unsupervised segmentation.
With zero scene-level supervision, our method performs well with a small gap in performance to the (weakly) supervised methods.

When the baselines are trained on \textbf{Z} but tested on \textbf{SO3}, the baselines do not show a large drop in performance, potentially because both \textbf{Z} and \textbf{SO3} scenes have no objects in contact with each other.
However, when tested on \textbf{Pile}, we see that baselines trained on \textbf{Z} or \textbf{SO3} perform differently, and are both worse than the ones trained on \textbf{Pile}, indicating a failure to generalize to clutter. Our method outperforms all the baselines that are not trained on \textbf{Pile} configurations.

We additionally generate 50 scenes each from three new and more difficult scene setups for testing on \textbf{Mugs} as shown in Fig.~\ref{fig:syn_results}.
In the \textbf{Tree} scenes mugs are hanging on a holder tree and distributed vertically.
The \textbf{Box} scenes include cubes to simulate objects that have not been seen during training.
In the \textbf{Shelf} scenes, the mugs are put on a shelf that is only visible from one side.
In all three of these scene setups, our method is able to outperform all baselines as the baselines are unable to generalize to configurations of mugs that are significantly outside of the training distribution.

\subsection{Results on real data}
\label{sec:exp_real}
\vspace{\vspaceaftersection}
We additionally evaluate the performance of our model on real data.
To the best of the authors' knowledge, there is no existing real dataset that contains interactable objects in diverse configurations for 3D instance segmentation. 
Therefore we collect a test set \textbf{Chairs and Mugs} that contains the reconstruction of 240 real scenes with object instance mask annotations to test our method in the real world. 

\paragraph{RealMugs:} As shown in Fig.~\ref{fig:real_results} we replicate the \textbf{Z, SO3, Pile} and \textbf{Tree} setups in the real world. 
The scene is captured by 4 calibrated realsense D455 cameras mounted on the corners of the table.
We further introduce two new setups that cover hard-to-simulate scenes. 
The \textbf{Others} setup contains random objects that a manipulator may encounter, including cloth, toys, paper bags, wires and tools. 
These objects are added into the tabletop scene, contacting and occluding the mugs.
The \textbf{Wild} setup contains crops of real-world indoor scans from labs, kitchens, and teaching buildings. Unlike~\cite{xu2022scene}, our scenes are not restricted to only showing the table and upright mugs, but include diverse configurations, backgrounds, and distractors. 
These scans are captured by an iPad with a lidar scanner. Since the \textbf{Z, SO3} and \textbf{Pile} setups are easy to simulate realistically, we only collect 10 scenes per setup. We collect 50 scenes for each of the \textbf{Tree}, \textbf{Others} and \textbf{Wild} setups.

We test all the baselines trained on the synthetic dataset and our method trained on ShapeNet directly on these real scenes.  Quantitative results are shown in~Tab.~\ref{tab:compact_mugs}. 
Since our simulator includes advanced techniques of active light simulation in the physical engine~\cite{xiang2020sapien} and we do not use the colours as input, the sim2real gap for the baseline methods is minimal in these controlled setups. 
We can draw similar conclusions for \textbf{Z}, \textbf{SO3}, \textbf{Pile} and \textbf{Tree} setups as with the synthetic experiments. 
Our method retains reasonable performance on \textbf{Others} scenes due to its awareness of the object shape. When tested on the \textbf{Wild} real-world scenes, our method also performs the best, which demonstrates the strong generalizability of our method.

\paragraph{RealChairs:} We also collect data of chairs in the real world following the \textbf{Z}, \textbf{SO3} and \textbf{Pile} setups. Although we have plenty of real-world scan datasets like~\cite{dai2017scannet}, 
 none of them include diverse configuration changes of chairs. 
 Therefore we collected and annotated a small test set with 20 scenes per setup as shown in Fig.~\ref{fig:real_results}. 
 We train the baselines in simulation and test them in the real world. 
 Additionally, we take all the baselines' official model weights from being trained on the real world ScanNet dataset~\cite{dai2017scannet} to evaluate on our test set. The results are shown in Tab.~\ref{tab:compact_real_chairs}. 
 There is a larger sim2real gap for the baselines with the chairs data than with the mugs data because of less realistic depth simulation and difficulties aligning the scale between Shapenet chairs and real-world chairs.
In contrast, the baselines trained on ScanNet work very well on the \textbf{Z} setup, which aligns best with the ScanNet dataset. However, they have a significant drop in performance when the testing distribution shifts to the \textbf{SO3} and \textbf{Pile} setups.
 In contrast, our method retains reasonable performance on real world chairs across different setups.
 We will release our Chairs and Mugs dataset to provide more opportunities to study robustness, generalizability and equivariance for scene and object understanding in the real world.

\begin{figure}[t!]
\centering
   \includegraphics[width=1.0\linewidth]{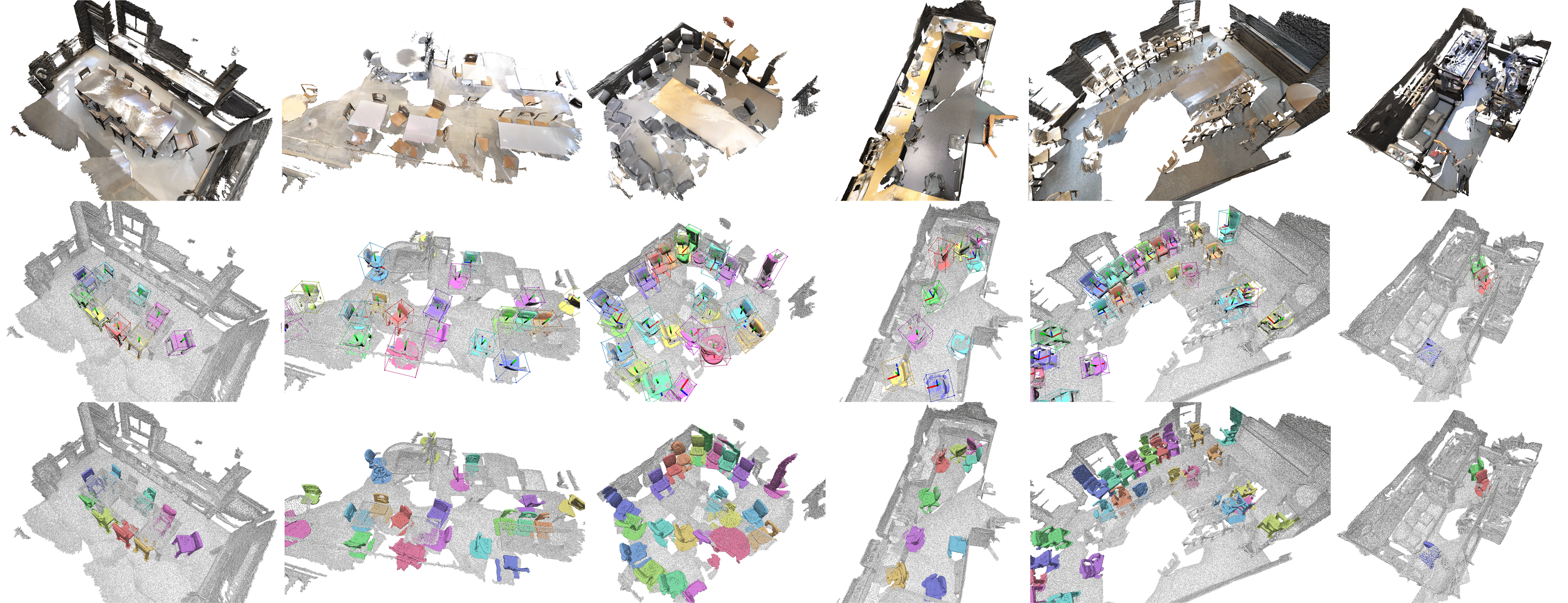}
    \\\vspace{-.7em}
    \caption{ScanNet qualitative results, the same format as Fig.~\ref{fig:syn_results}}.
    \label{fig:scannet}
\end{figure}

We also show our effectiveness on ScanNet~\cite{dai2017scannet}, where the indoor scene scan can span the whole room. 
Qualitative results are shown in Fig.~\ref{fig:scannet} and the AP metrics for the chairs category of our method on the validation/test set are $AP=24.6/20.2$, $AP50=50.8/39.0$ and $AP25=61.3/48.3$ where in comparison the weakly supervised method CSC~\cite{hou2021exploring} trained on 200 points labels achieves $AP50=62.9/61.1$ (See Suppl. for a table). One main reason for our performance drop on ScanNet is that ScanNet has many  partially observed chairs, which are hard to be recognized via shape.
We leave future explorations to fill this gap between our unsupervised method and the (weakly) supervised ones.

\vspace{-.5em}
\subsection{Ablations}
\vspace{-.5em}
\setlength{\columnsep}{5pt}
\begin{wraptable}{r}{0pt}
\raisebox{0pt}[\dimexpr\height-0.2\baselineskip\relax]{
\centering
\scalebox{0.8}{
\begin{tabular}{c|ccc}
\toprule
Ablation   & AP             & AP50           & AP25           \\ \midrule
Full       & \textbf{0.601} & \textbf{0.789} & \textbf{0.807} \\
No Phase 2 & 0.548          & 0.740          & 0.756          \\
No Normal  & 0.203          & 0.302          & 0.337          \\ \bottomrule
\end{tabular}
}
}
\caption{Ablations}
\label{tab:abl}
\end{wraptable}

We verify the effectiveness of our design by removing the phase-2 joint iterations and removing the usage of the normals. 
When not using phase-2 (Sec.~\ref{sec:full_alg}), we let the phase-1 independent iterations run more steps to keep the total number of iterations constant. 
When removing the normals, all error computing, assignment weight updating, and confidence scoring will only take the distance error term in to account while ignoring the normal term. 
We compare our full model with the ablated models on SynKit \textbf{Pile} setups. The quantitative results are shown in Tab.~\ref{tab:abl}, which illustrates that both components contribute to our model's overall performance. More ablation studies can be found in our supplementary.

\vspace{\vspacebeforesection}
\section{Conclusions}
\vspace{\vspaceaftersection}
We present EFEM, a method for 3D instance segmentation that is trained only on shape datasets.  Without requiring any real or simulated scene data, our method can generalize to complex, real world scenes better than existing methods that also require more supervision.

\paragraph{Limitations and future work}
Although we show an encouraging step towards unsupervised 3D instance segmentation by generalizing knowledge directly from a shape collection to a scene, several weaknesses remain. First, our method has a performance drop when the object is significantly occluded.
This is due to our method recognizing objects only via their shape, rather than also reasoning about color or other features. 
Second our current running speed is slow ($\sim$1min per scene) due to the large number of proposals the initialization requires.

\paragraph{Acknowledgements} The authors appreciate the support of the following grants: NSF FRR 2220868, NSF IIS-RI 2212433, NSF TRIPODS 1934960,  ARL DCIST CRA W911NF-17-2-0181, ARO MURI W911NF-20-1-0080, ONR N00014-22-1-2677 awarded to UPenn; Toyota University 2.0 program and Vannevar Bush Faculty Fellowship awarded to Stanford University.

{\small
\bibliographystyle{ieee_fullname}
\bibliography{reference.bib}
}

\end{document}